\documentclass{article}
\usepackage{spconf,amsmath,graphicx, url, tablefootnote}

\graphicspath{{figures/}} 

\title{Cantonese Automatic Speech Recognition Using Transfer Learning from Mandarin}
\twoauthors
 {Bryan Li\textsuperscript{1}, Xinyue Wang\textsuperscript{1}}
	{Columbia University\textsuperscript{1}\\New York, NY\\
	\lbrack bl2557, xw2368\rbrack@columbia.edu}
 {Homayoon Beigi\textsuperscript{1,2}}
	{Recognition Technologies, Inc.\textsuperscript{2}\\White Plains, NY\\
	beigi@recotechnologies.com}
\begin{document}
%
\maketitle
\thispagestyle{plain}
\pagestyle{plain}
\begin{abstract}
We propose a system to develop a basic automatic speech recognizer(ASR) for Cantonese, a low-resource language, through transfer learning of Mandarin, a high-resource language. We take a time-delayed neural network trained on Mandarin, and perform weight transfer of several layers to a newly initialized model for Cantonese. We experiment with the number of layers transferred, their learning rates, and pretraining i-vectors. Key findings are that this approach allows for quicker training time with less data. We find that for every epoch, log-probability is smaller for transfer learning models compared to a Cantonese-only model. The transfer learning models show slight improvement in CER.
\end{abstract}
\section{Introduction}
\label{sec:intro}
Much of the work done thus far on automatic speech recognition has focused on a subset of high-resource languages, such as English, Spanish and Mandarin Chinese -- these languages are fortunate to have expert-built dictionaries, part-of-speech taggers, language models, as well as large-amount of high-quality voices recorded in professional studios. All these allow for development of state-of-art ASR systems. However, there are approximately 6500 languages in the world, many of which spoken by millions of people, yet have no such resources and are little studied for speech recognition. Interest has recently been taken in working with low-resource languages. This work both generalizes and strengthens language and speech modeling technologies, and opens up speech technologies to a new cohort of speakers.

Cantonese is a variety of Yue Chinese spoken by over 100 million people in Southeastern China. Most speakers of Cantonese are fluent in Mandarin, the national language, and one at the forefront of speech research. As a result, there is a general lack of interest in Cantonese such that it is a low-resource language. This is further complicated by its lack of a standardized written form\footnote{Hong Kong Cantonese is, but we consider mainland Cantonese only.} because Cantonese is a spoken-dialect, which has made it difficult for quality textual and audio corpora to be collected.

Though varieties of Chinese are often thought of as merely dialects, they are not mutually intelligible and just as distinct as family like Romance languages. Some key differences between Cantonese and Mandarin are: 6 vs. 4 tones, presence of final stops, lack of initial retroflexes, and colloquial lexicon. Still, the languages share a common ancestor and many similarities. Thus, we aim to leverage the myriad of resources available in Mandarin Chinese to train a state-of-the-art Mandarin ASR system. From here, we apply transfer learning to initialize parameters of a Cantonese ASR system, training further on a limited Cantonese dataset.

\section{Related Work}
\label{sec:related-work}
\subsection{Transfer Learning}
\label{ssec:transfer-learning}
Transfer learning is a vital technique that generalizes models trained for one task to other tasks \cite{DBLP:journals/corr/WangZ15d}. In speech processing, some common patterns represented by features such as MFCCs and pitches are shared across languages, especially for similar/geographically closely located languages such as Cantonese and Mandarin. Neural network models have many parameters, and as such depend on large volumes of data to learn the patterns of a problem. Furthermore, they are not robust to low-quality training data with errors. Both issues are prevalent when working with low-resource languages. Transfer learning helps to alleviate these issues, since we are able to use well-maintained corpora from high-resource languages. A key idea is that the features learned by deep neural network
models are more language-independent in earlier layers, and more language-dependent in later layers~\cite{DBLP:journals/corr/WangZ15d}.

Zoph et al.~\cite{DBLP:journals/corr/abs-1809-00357} show the effectiveness of applying transfer learning on different languages  in the domain of machine translation. Initializing from parameters of a parent model in French-English and training child models in low-resource languages such as Turkish and Urdu, they see significant increases in BLEU scores. They find better gains when applying transfer learning to related languages, such as French and Spanish, to unrelated languages, such as French and German.

Kunze et al.~\cite{kunze2017transfer} develop a German ASR system using the transfer learning approach of \textit{model adaptation}~\cite{DBLP:journals/corr/WangZ15d}. This involves first training a model on one language (in this case English), and then retrain all or parts of it on a different language (German). They use an 11-layer convolutional neural network (CNN), and experiment with fixing \textit{k} layers. They find that such weight initialization allows for much faster training without a decrease in quality. Freezing more layers (higher \textit{k}) decreases training time, but the best-performing model allows all layers to learn (\textit{k}=0). Our report seeks to experiment similarly, but with a time-delayed neural network (TDNN) on Mandarin to Cantonese.

\subsection{State-of-Art for Cantonese ASR}
\label{ssec:cantonese-asr}
We primary consider the BABEL Cantonese dataset, further described in Section \ref{sec:datasets}. 
Karafiat et al. \cite{but-babel-cantonese} explore feature extraction using a 6-layer stacked bottleneck neural network. These outputs are used as features for a GMM-HMM recognition system. They concatenate these features with fundamental frequency f0, reasoning that pitch is important to Cantonese, a tonal language. Further, they apply region-dependent transforms, a speech-specific GMM optimization method \cite{but-babel-cantonese}, to all these features, and arrive at a character-error rate (CER) of 42.4\%.

Further work is done by Du et al. \cite{7472833-machine-translation}, who note an inherent difficulty: because there is no standardized form of Cantonese, it is difficult to perform web-scraping tasks in isolating Cantonese text. Therefore, they combine the BABEL Cantonese transcriptions with data-augmented Cantonese text machine translated from transcriptions of Mandarin telephone conversations. Using 88 features including pitch, Mel-PLP, and TRAP-DCT input into a bottleneck DNN, they are able to improve the CER to 40.2\%.

\section{Datasets Used}
\label{sec:datasets}

The Cantonese dataset is IARPA Babel Cantonese \cite{babel-dataset}. This includes 215 hours of Cantonese conversational and scripted telephone speech, along with corresponding transcripts. Speakers represent dialects from the Chinese provinces of Guangdong and Guangxi. The 952 speakers range from ages 16 to 67, and the male-female ratio is 48\%:52\%. Audio was collected from mobile and landline telephones, and from environments such as the street, at home, and inside a vehicle. We train and test on only the conversational data (75\% of all data), as done by all prior work with this dataset.

The Mandarin dataset is AISHELL2~\cite{DBLP:journals/corr/abs-1808-10583}, which consists of 1000 hours of clean read-speech data [3], along with corresponding transcripts. The 1991 speakers range from ages 11 to 40 that speak on 12 topics such as voice control, news, and  technology. The male-female ratio is 40\%:60\%. Audio was collected from three sources. From left to right: iOS device, a high-fidelity microphone, and an Android device. We use only the iOS-recorded data, as it produced the best results for the Mandarin model. 

\subsection{Preprocessing}
\label{sec:preprocessing}
The Mandarin data has a sample rate of 16 kHz, whereas the Cantonese data has a sample rate of 8 kHz. To be able work with the same extracted features, we downsample Mandarin audio to 16 kHz. However, the audio quality of BABEL is still worse, and about half of the audio is silence.

All data preprocessing, including resampling and conversion to wav format, was done with \texttt{sox}. Speed perturbation of each file (90\%, 100\%, 110\%) is performed to augment the training data, and volume perturbation is performed to make models more invariant to test data volume.

\subsection{Feature Extraction}
\label{ssec:features}
The GMM models are trained with 13-dimensional Mel-frequency cepstral coefficients (MFCCs) and 3 pitch features. Note that Babel originally used Perceptual Linear Prediction (PLP) features. The TDNN models are trained with 39-dimensional MFCCs and 4 pitch features. They also include for each frame-wise input a 100-dimensional i-vector. Two extractors are trained, one for each of Cantonese and Mandarin. We extract Cantonese i-vectors using one or the other for different models.

\section{Model Architecture}
\label{sec:architecture}
Our model is implemented in Kaldi, and follows a fairly standard Kaldi pipeline. Our recipe is written with reference to existing ones --TEDLIUM, BABEL, AISHELL2.

\subsection{Language Model}
\label{ssec:lm}
Written Chinese does not have spaces between words, so ASR systems require sophisticated word segmentation. AISHELL2 \cite{DBLP:journals/corr/abs-1808-10583} includes a open-source dictionary called DaCiDian, which maps Chinese words in two layers, first from word to PinYin syllables~\cite{duanmu2007phonology}
, and second from PinYin to phoneme. Word segmentation is done with Jieba~\cite{sun2012jieba}, which uses a prefix-tree based search approach, and the Viterbi algorithm for out-of-vocabulary (OOV) words.

In parallel with AISHELL, we present a Cantonese dictionary called DaaiCiDin, which maps Cantonese words to Yale romanization system syllables, then maps these to X-SAMPA phonemes. Its vocabulary is compiled from training set transcripts of Babel Cantonese.

The statistical language model is implemented in SRILM, and consists of n-grams of size 2, 3, and 4 using both Kneser-Ney and Good-Turing smoothing. Silence and OOV words are also accounted for.

\subsection{Acoustic Model}
\label{ssec:acoustic}
The acoustic model consists of two stages: a GMM-HMM model, and a TDNN-LSTM model.

\subsubsection{GMM}

\begin{figure}[t]
\begin{center}
   \includegraphics[width=1\linewidth]{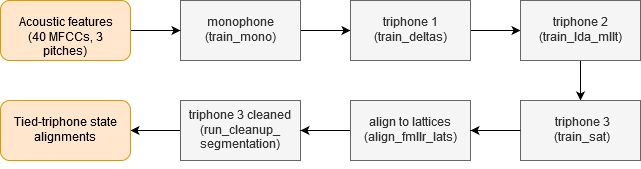}
\end{center}
   \caption{GMM-HMM acoustic model}
\label{fig:gmm}
\end{figure}

\label{sssec:gmm}
The Gaussian Mixture Model is shown in Figure \ref{fig:gmm}. It is trained with an input dimension of 16 (13 MFCCs plus 3 pitches). A monophone model is trained as the starting point for the triphone models. Then, a larger triphone model is trained using delta features. To replace the delta features, linear discriminant analysis (LDA) is applied on stack of frames, and MLLT-based global transform is estimated iteratively. Next comes the Speaker Adaptive Training (SAT) stage, whose outputs are aligned to lattices. Finally, we run Kaldi clean-up scripts so all audio in the training data matches the transcripts. The GMM model outputs the tied-triphone state alignments.

\subsubsection{TDNN}
\label{sssec:tdnn}
\begin{figure}[t]
\begin{center}
   \includegraphics[width=1\linewidth]{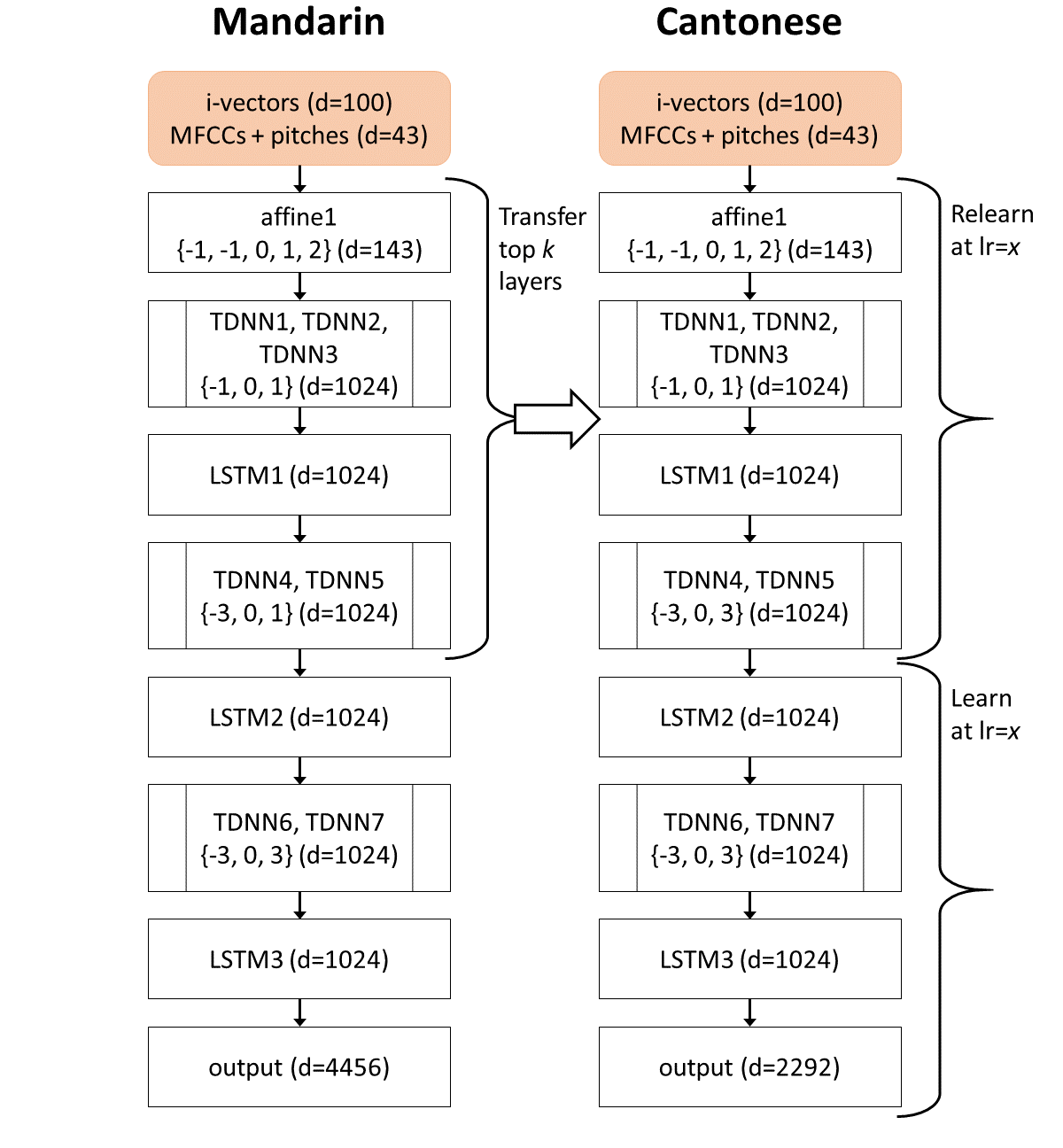}
\end{center}
   \caption{TDNN-LSTM model + transfer learning from Mandarin to Cantonese}
\label{fig:architecture}
\end{figure}

The TDNN-LSTM model consists of time-delayed neural network ~\cite{Peddinti2015ATD} and long-short term memory layers. It alternates between the layer types as shown in Figure \ref{fig:architecture}.

We first train a TDNN on AISHELL2. The objective function is lattice-free maximum mutual information (LF-MMI)~\cite{povey2016purely}. The network also uses sub-sampling as described in \cite{Peddinti2015ATD} and is not fully connected -- the input of each hidden layer is a frame-wise spliced output of its preceding layer \cite{DBLP:journals/corr/abs-1808-10583}. The input to the network is high resolution MFCC with cepstral normalization plus pitches, making the dimension 43. In addition, a 100-dimensional i-vector is attached to each frame-wise input using the extractor trained on AISHELL2 to encode speaker dependent information. i-vectors use as features high-resolution MFCCs, pitches, and the GMM-HMM tied-triphone state alignments.

In the transfer learning stage, we adapt the TDNN trained for Mandarin as described above to the Cantonese language. We initialize the weights of the first \textit{k} layers of the TDNN to be those of the Mandarin model, and set their learning rate = \textit{x}. Then we randomly initialize the weights of remaining layers, and set their learning rate = 1.0. i-vectors are obtained using either the AISHELL2 or BABEL Cantonese i-vector extractor depending on our experimental configuration.

\section{Experiments}
\label{sec:experiments}
We perform several experiments to explore the space of transfer learning. All our experiments use the model architecture as described above, and modify the learning rate of the transferred layers when training the Cantonese model.  The learning rates we used are 0, 0.25, and 1.0. These are relative to the learning rate of the final block, which is always 1.0. lr=0 corresponds to fixing transferred learning weights, whereas 0.25 and 1.0 allow the model to adjust them as necessary.

For lr=0.25, we additionally train a model that uses the i-vector extractor of AISHELL2 on the BABEL Cantonese dev set. For comparison purposes, we also train two baseline Cantonese models: one using the TDNN recipe from BABEL, and one using the AISHELL2 architecture without transfer learning (all weights initialized to zero).

The same hyperparameters are used to train all models: number of epochs=4, initial learning rate=0.001, final learning rate=0.0001, minibatch size=128, cross-entropy regularization=0.1, and frames per example=150, 110, and 90. The dropout schedule is 0, 30\% after 50\% of the data is seen, and 0 for the last layer. 

\section{Results and Discussion}
In this section we report experimental results of transfer learning. First, we compare model performance in terms of log-probability and training time. Second, we compare ASR performance in terms of character and word error rates.

\subsection{Metrics Used}
\textbf{Word error rate (WER)} is the standard metric used for speech recognition systems, and is given by:
\[WER=\frac{S+D+I}{N}\]
S, D, and I are the number of substitutions, deletions, and insertions respectively. N is the total number of words in the reference transcription. WER ranges from 0-300\%.

\textbf{Character error rate (CER)} follows the same formula, but with characters as the unit. In written Chinese, the majority of words are made of two characters. Following prior work, we primarily report CER results.

\subsection{Model Loss}
\label{ssec: diff-lr}
\begin{figure}[t]
\begin{center}
   \includegraphics[width=1\linewidth]{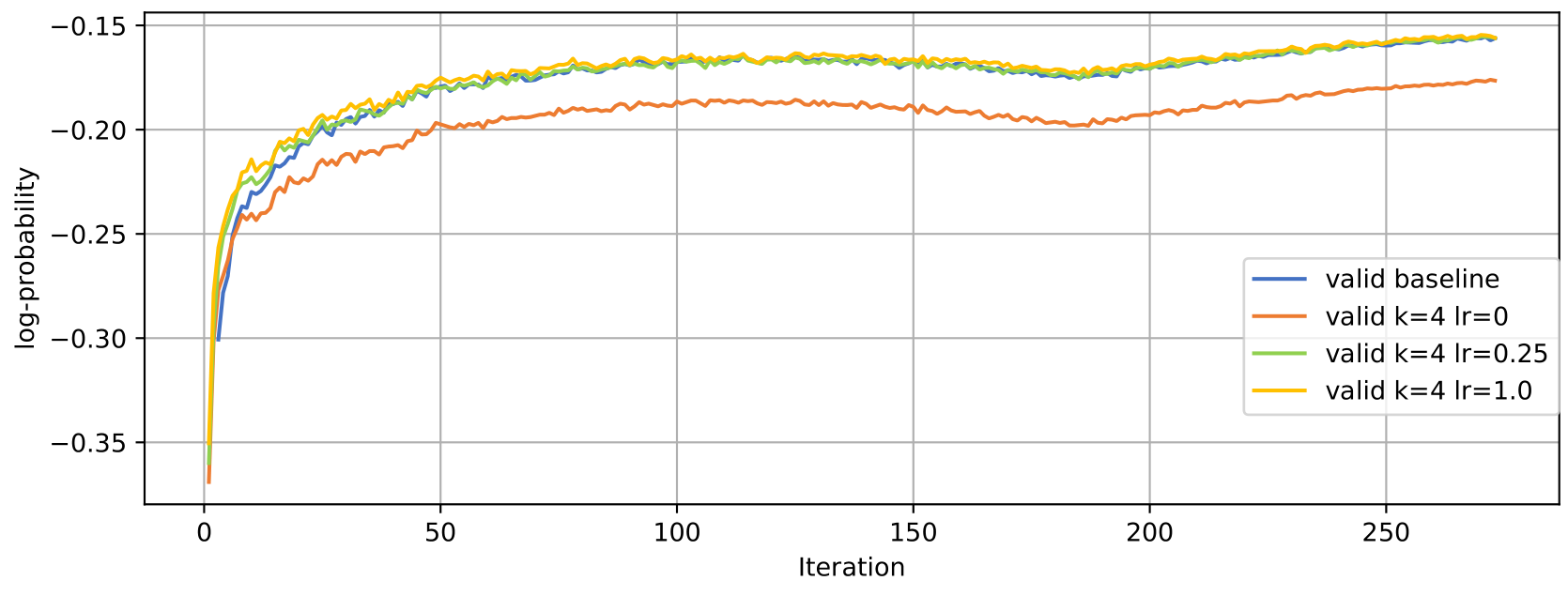}
\end{center}
\caption{Plot of log-probability vs iteration for output}
\label{fig:2}
\end{figure}

Figure \ref{fig:2} compares the the log-probability (log-prob) over time for different models. The results show that log-prob (of cross-entropy loss) is always higher for transfer learned models, for both train and validation sets. Higher learning rates correspond to higher log-prob. 

These findings are consistent with those of  Kunze et al.~\cite{kunze2017transfer}. Because the languages are not the same (for example, Cantonese uses long vowels instead of diphthongs~\cite{manda_sampa, canto_sampa}), it is better to allow the earlier layer weights to adjust themselves than to fix them.

\subsection{Reduced Computing Time}
\label{ssec: reduced-comput}
\begin{figure}[t]
\begin{center}
   \includegraphics[width=1\linewidth]{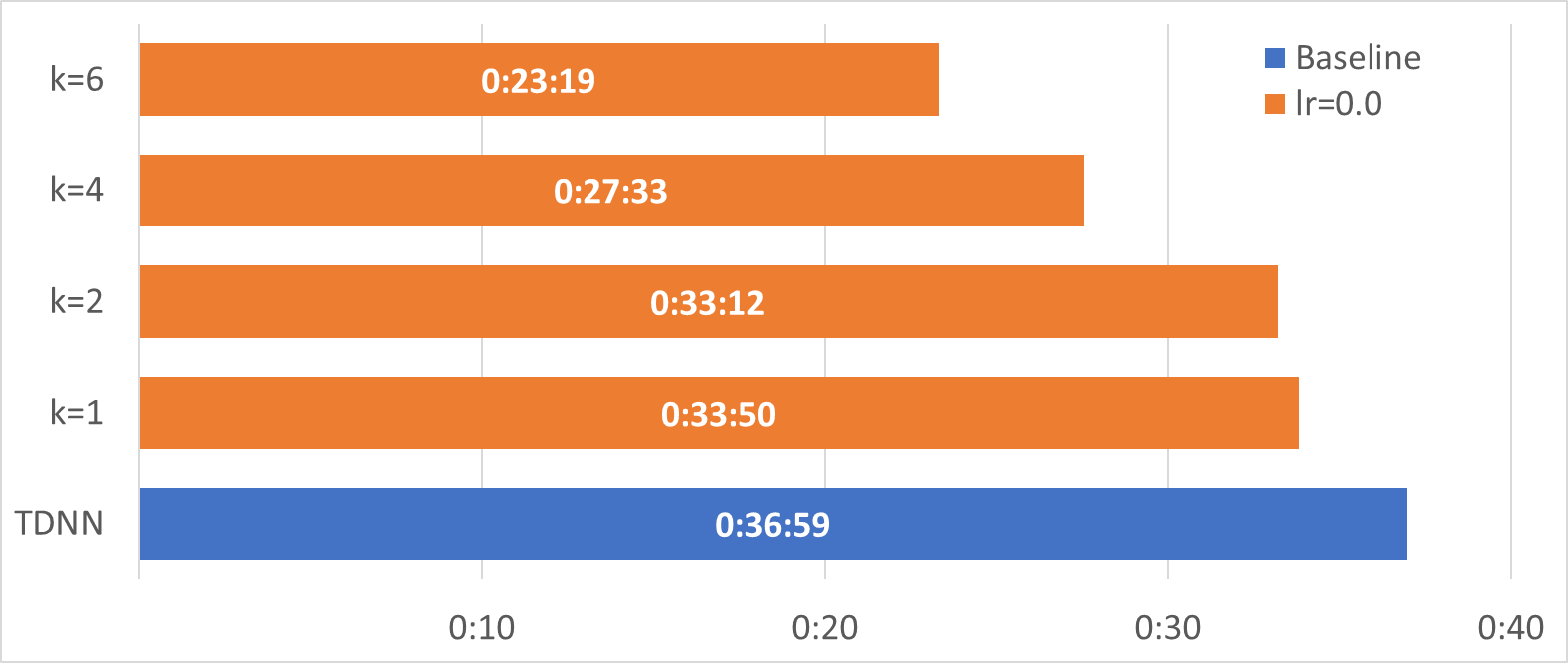}
\end{center}
\caption{Comparison of training time in minutes per iterations for different learning rates}
\label{fig:3}
\end{figure}

Given that languages share common features, earlier layers in the model should contain common features that can be transferred to that of a different language. Thus, we can fix the earlier layers of the original Mandarin model, and this decreases the average training time per iteration because the network will need to back-propagate though fewer layers. Train times depending on \textit{k} are shown in Figure \ref{fig:3}. Note that for all models lr>0, train time is equivalent to the baseline train time.

However, from our discussion in section \ref{ssec: diff-lr}, loss is always smaller at any point for non-zero learning rate. Thus, we figure that in terms of achieving small loss, there is no reason to freeze the early layers. Nevertheless, when we compare the log-probability of the transferred model with the baseline Cantonese model trained from scratch, we see the transferred model starts off with a much lower loss and thus is able to achieve the same loss with shorter time. Therefore, it is beneficial in terms of computing time to transfer learn with a good weight initialization.

\subsection{Character and Word Error Rates}

\begin{figure}[t]
\begin{center}
   \includegraphics[width=1\linewidth]{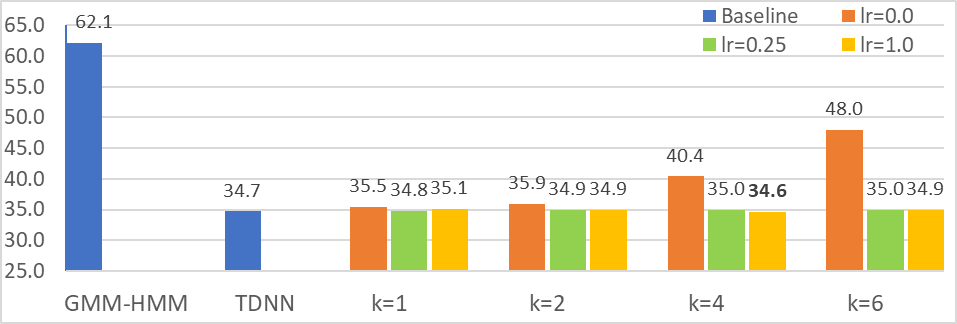}
\end{center}
\caption{Character Error Rate of transfer-learned models vs baseline}
\label{fig:cer}
\end{figure}

Figure \ref{fig:cer} shows the speech recognition performance of the various systems on the development set. For lr=0.0, the more transferred layers, the worse results are -- this makes sense because of the distance between the languages. We should still do some learning. For lr=0.25 and 1.0, results are similar across the board. We see that our best model, k=4 and lr=1.0, gives a slight increase of 0.1\% CER vs the baseline. 

We do not report results for the experiments with pre-training i-vectors on Mandarin, and transferring to Cantonese, as they do not improve performance. We suspect that weight transfer could help here, as with the TDNN-LSTM, but leave that to future work.

\section{Future Work}
\label{set:future}

For future work, we have several approaches in mind. First, despite downsampling AISHELL-2 to 8 kHz to match BABEL Cantonese, the latter still has far worse audio quality. Therefore, we will try adding acoustic noise Mandarin, and training the model to be transferred on that. Second, we aim to explore the language model more. The LMs are only trained on acoustic transcripts, and we can add more data. Also, prior linguistic work has shown that correspondence can be made between Cantonese and Mandarin phonemes (and to a lesser extent, tones), so we can create a mapping between phonemes. Most written Chinese characters are words in both Cantonese and Mandarin, so that would help with mapping. Finally, fine-tuning the AISHELL-2 i-vectors after pre-training would be a good experiment.

\section{Conclusion}
\label{sec:conclusion}
We propose a system to develop a basic ASR for Cantonese through transfer learning of Mandarin. We first trained a TDNN for Mandarin using large amounts of data from AISHELL2, and adapted the model to Cantonese by transferring the weights and adjusting the relative-learning rates of earlier and later layers. We conducted multiple experiments setting lr = 0, 0.25 and 1 and show that log-probability is always higher for transfer learned models. In addition, we showed reduced computation time for freezing earlier.

We achieve a new state-of-the-art on BABEL Cantonese (CER=34.6\%) by transferring up to TDNN4 and using lr=1.0. Results are not as promising as prior work has been, but we believe our work is still as step in the right direction. We hope to make the transfer learning methods more informed than simple model adaptation.


\bibliographystyle{IEEEtran}
\bibliography{refs}

\end{document}